\pdfoutput=1

\documentclass[11pt]{article}

\usepackage[preprint]{acl}

\usepackage{times}
\usepackage{latexsym,enumitem, comment}

\usepackage[T1]{fontenc}

\usepackage[utf8]{inputenc}

\usepackage{microtype}

\usepackage{inconsolata}

\usepackage{graphicx}

\title{Towards Safer Online Spaces: Simulating and Assessing Intervention Strategies for Eating Disorder Discussions}

\author{
 \textbf{Louis Penafiel\textsuperscript{1}},
 \textbf{Hsien-Te Kao\textsuperscript{1}},
 \textbf{Isabel Erickson\textsuperscript{1}},
 \textbf{David Chu\textsuperscript{2}},
\\
 \textbf{Robert McCormack\textsuperscript{1}},
 \textbf{Kristina Lerman\textsuperscript{2}},
 \textbf{Svitlana Volkova\textsuperscript{1}},
\\
\\
 \textsuperscript{1}Aptima, Inc.,
 \textsuperscript{2}USC Information Sciences Institute
\\
 \small{
   \textbf{Correspondence:} [lpenafiel, hkao, ierickson, rmccormack, svolkova]@aptima.com, [dchu, lerman]@isi.edu
 }
}

\begin{document}
\maketitle
\begin{abstract}
Eating disorders are complex mental health conditions that affect millions of people around the world. Effective interventions on social media platforms are crucial, yet testing strategies in situ can be risky. We present a novel LLM-driven experimental testbed for simulating and assessing intervention strategies in ED-related discussions. Our framework generates synthetic conversations across multiple platforms, models, and ED-related topics, allowing for controlled experimentation with diverse intervention approaches. We analyze the impact of various intervention strategies on conversation dynamics across four dimensions: intervention type, generative model, social media platform, and ED-related community/topic. We employ cognitive domain analysis metrics, including sentiment, emotions, etc., to evaluate the effectiveness of interventions. Our findings reveal that civility-focused interventions consistently improve positive sentiment and emotional tone across all dimensions, while insight-resetting approaches tend to increase negative emotions. We also uncover significant biases in LLM-generated conversations, with cognitive metrics varying notably between models (Claude-3 Haiku $>$ Mistral $>$ GPT-3.5-turbo $>$ LLaMA3) and even between versions of the same model. These variations highlight the importance of model selection in simulating realistic discussions related to ED. Our work provides valuable information on the complex dynamics of ED-related discussions and the effectiveness of various intervention strategies.

\end{abstract}

\section{Introduction}
Eating disorders (ED) are complex mental health conditions that affect more than 24 million people in the United States, characterized by unhealthy thoughts and behaviors related to food and body image~\cite{Hoeken2020,psy2023us,mayo2023us,psy2023us,nimh2023us}. The sensitive nature of ED communities necessitates highly effective interventions~\cite{whitehouse2023us,hhs2023us,senate2023us,ped2023us}. Understanding when to intervene and the impact of interventions is crucial, as is identifying the most effective types of interventions across various social media platforms and communities.
Currently, testing intervention methods is limited to in situ deployment on social media platforms~\cite{twitter2023,reddit2024}, which can be risky without a thorough understanding of their impact. To address this challenge, we have developed an LLM-driven experimental testbed that generates social media conversations on relevant topics across platforms. This innovative approach allows for controlled experimentation with diverse intervention strategies, enabling the measurement of their effects in vitro. Our research aims to:
\begin{itemize}[noitemsep, nolistsep]
\itemsep0em
\item Evaluate the efficacy of various intervention strategies in a controlled environment.
\item Analyze differences in intervention effects across social platforms and communities.
\item Quantify variations between generative models used in synthesizing conversations.
\end{itemize}

The operational impact of this research extends beyond civilian contexts into military settings, where maintaining personnel health and readiness is paramount \cite{bodell2014consequences,antczak2008diagnosed,bodell2014consequences}. Effective ED interventions can  improve force readiness and operational effectiveness, particularly in high-stress environments where disordered eating behaviors may be exacerbated \cite{smith2020prevalence,mitchell2023impact,bauman2024incidence}. By providing a controlled testbed for intervention strategies, our approach enables the development of more targeted and efficient support systems for service members, potentially reducing attrition rates and enhancing overall unit performance.

\section{Methodology}
\subsection{LLM-driven Experimental Testbed}
We employ a novel LLM-driven experimental testbed to simulate dyadic conversations across various social media platforms (Reddit, Twitter and ED Forums), ED communities (automatically identified on each platform~\cite{Kao_CHI_24}), and language models (Claude-3 Haiku, Mistral, GPT3.5-turbo, LLaMA3). This testbed enables the injection of diverse mediation strategies at controlled points within the generated conversations. Each simulated dialogue undergoes comprehensive cognitive domain analysis, including metrics such as sentiment, emotion detection, and empathy assessment.
Figure 1 illustrates the conversation generation workflow, which encompasses prompt design, mediation strategy implementation, conversation synthesis, and subsequent analysis. This systematic approach allows for the generation and evaluation of thousands of representative conversations, providing a robust framework for assessing intervention effectiveness without the ethical concerns and potential risks associated with direct human interaction.
Our testbed offers researchers and practitioners a safe, controlled environment to develop, implement, and analyze intervention experiments at scale. By simulating a wide range of scenarios and interventions, we can identify the most promising strategies for supporting individuals in ED-related online discussions before deploying them in real-world settings.

\subsection{Datasets}
Our study utilizes diverse datasets to capture a comprehensive view of eating disorder (ED) discussions across multiple platforms. We draw upon the work of ~\citeauthor{twitter2023} and ~\citeauthor{reddit2024} for Twitter and Reddit discussions, respectively. To complement these sources, we incorporate data from ED Support Forum, a traditional online forum that provides a safe space for individuals affected by EDs.
ED Support Forum is particularly valuable for our research as it hosts discussions on sensitive topics that are often censored or prohibited on mainstream social media platforms. This unique dataset offers insights into conversations that are not readily available elsewhere, providing a more complete picture of ED-related discourse.
To ensure ethical data collection, we leverage ED Support Forum's RSS feed, carefully scraping content while rigorously removing all personally identifiable information (PII) and other sensitive details. This approach allows us to analyze authentic ED-related discussions while maintaining user privacy and adhering to ethical research standards.

\subsection{Prompting Strategies}

To develop LLM prompts, we employ a discussion-focus detection framework to extract clusters of eating disorder-related discussions from Twitter, ED Support Forum, and Reddit~\cite{Kao_CHI_24}. The algorithm segments posts into subject-verb-object tuples, de-duplicates and filters chunks, and embeds them using SentenceTransformer~\cite{reimers2019sentence}. These embeddings are iteratively clustered, maximizing separation and coherence. This process results in a final set of 7 distinct discussion clusters per platform.

For Twitter, these clusters include discussions on keto-friendly recipes, eating disorder-related Twitter posts, periods on the Keto diet, weight loss achievements (in kilograms and pounds), communities related to eating disorders, and discussions about personal weight measurements (in kilograms and pounds). In the ED Support Forum, the groups encompass shared experiences about eating disorders, transitions between diet phases, sharing food recipes, discussions about calorie intake (specific amounts), weight measurement updates and weight loss achievements (in pounds). Reddit discussions focus on reading and responding to comments, calorie intake discussions (specific amounts), transition phases between dieting, personal weight measurements, sharing food recipes, weight loss achievements (in pounds), and estimating food calories. Several discussion topics are common across all three platforms, reflecting shared interests and concerns among users. These identified discussion clusters serve as the foundation for constructing targeted prompts for our simulation. Below is an example prompt and a single turn of a generated dialog: {\it The following is a transcript of an argument on posting about Keto diet period between Jane and John on Twitter. JANE: I am on the keto diet for several weeks. JOHN: You should not continue this unhealthy diet. }

\subsection{Intervention Strategies}
Our study incorporates a diverse array of mediation strategies, informed by principles of civil discourse and clinical expertise. To ensure consistency in our experimental design, all interventions are introduced immediately following the initial prompt, maintaining a uniform timing across different strategies. These interventions span a spectrum of intensity, ranging from subtle guidance to more assertive moderation. The mediation strategies, ordered from least to most intensive, include: no mediation (control), supportive recommendations, Socratic questioning to promote self-reflection, factual health information provision, modeling civil discourse, reminders about community guidelines, and warnings about potential account restrictions. Each strategy is associated with specific intervention language, carefully crafted to reflect its intended purpose and intensity level. This graduated approach allows us to assess the effectiveness of various interventions in managing and improving online discussions related to eating disorders.
\begin{itemize}[noitemsep,nolistsep]
    \item \textbf{No Mediation} Conversations without any intervention. The baseline to compare with intervention strategies. 
    \item \textbf{Support} ``I might not understand, but if you need someone to talk to I will listen and help as much as I can.''
    \item \textbf{Reset Insight} ``Let's take a step back. How do these comments provide additional insight about the topic?''
    \item \textbf{Health Advice} ``If you are concerned about a potentially serious medical issue, please contact your local health-line, doctor, or go to the nearest emergency room or urgent care.''
    \item \textbf{Modeling Civility} ``Discussion is most productive when the language is civil. Let’s keep it respectful. Thanks for your assistance.''
    \item \textbf{Rules Civility} ``Hey, I need to cut in. We need everyone to follow the rules for this discussion for this space to work. Please review the rules if you are unfamiliar and then follow them.''
    \item \textbf{Informing Banned} ``Note violations of community guidelines will lead to bans.''
\end{itemize}

\begin{figure*}[hbt!]
\centering
  \includegraphics[width=0.85\textwidth]{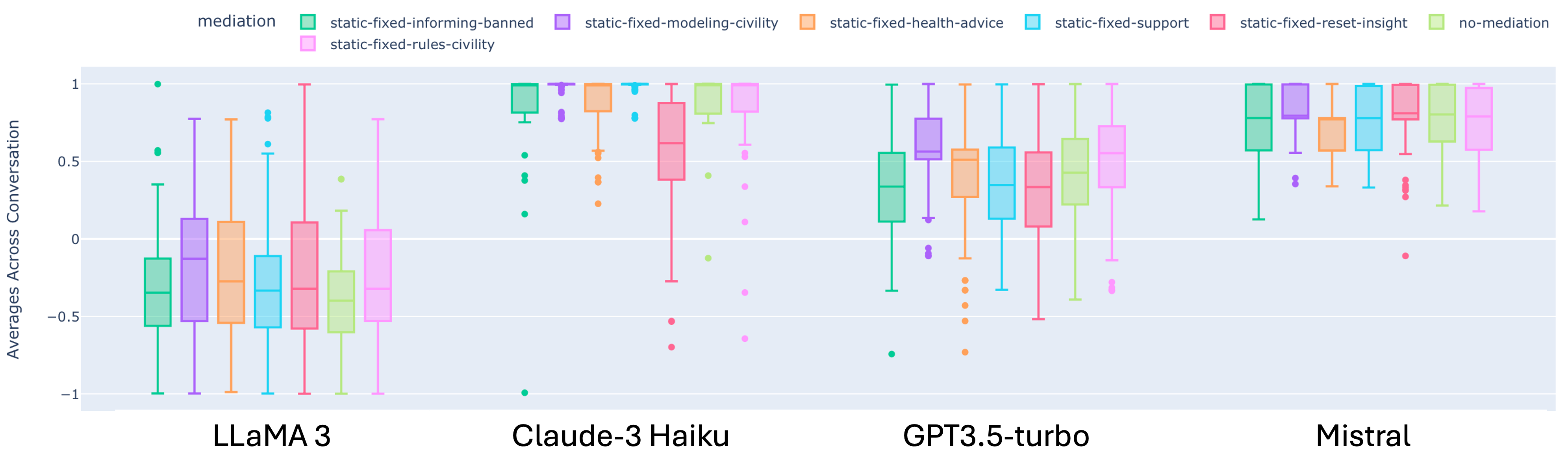}
  \caption{{\bf Cross-model analysis:} Comparing intervention effectiveness across different mediation strategies using average sentiment scores over simulated conversations by each LLM (colors represents  intervention strategies).}
  \label{fig:cross-model-sentiment}
\end{figure*}

\begin{table*}[t!]
\centering
\small
\begin{tabular}{| l | r | r | l | l |} 
 \hline
 LLM & $Med(s_{none})$ & $\overline{Med(s)_{diff}}$ & $max(Med(s)_{diff})$ & $min(Med(s)_{diff})$ \\ [0.5ex] 
 \hline
 Mistral & 0.613 & 0.093 & 0.173, support & -0.054, reset insight \\ 
 \hline
 LLaMA3 & -0.399 & 0.111 & 0.271, modeling civility & 0.052, informing banned\\
 \hline
 Claude-3 Haiku & 0.999 & -0.067 & 0.000, modeling civility & -0.382, reset insight \\
 \hline
 GPT3.5-turbo & 0.437 & 0.014 & 0.137, modeling civility & -0.092, reset insight\\ 
\hline
\end{tabular}
\caption{Median, min and max sentiment scores across LLMs with most and least effective interventions.}
\vspace{-0.3cm}
\label{table:cross-model-sentiment}
\end{table*}

\subsection{Generative Models}
We conducted experiments with the LLMs listed below: GPT-3.5-turbo-0125 (current version of GPT-3.5-turbo), GPT-3.5-turbo-1106, GPT-3.5-turbo-0613, Claude-3-Haiku-0307, Mistral, LLaMA2, and LLaMA3. These sets of models were chosen because of either their open-sourced availability, e.g., Mistral and Llama3, or their non-prohibitive cost structure, e.g., GPT-3.5 or Claude-3-Haiku. Lastly, the versions of the models were chosen, based on availability, e.g., GPT-3.5 has those available model versions, while Claude does not have older versions available.

\subsection{Measuring the Effect of Interventions using Cognitive Domain Analytics}
To measure the effects of interventions we employ seven cognitive domain analytics - sentiment \cite{sentiment:huggingface}, toxicity \cite{Detoxify}, empathy (emotions, intent) \cite{lee2022does}, emotions \cite{DBLP:journals/corr/abs-1810-04805}, moral values \cite{garten2016morality}, connotation frame analysis \cite{rashkin2015connotation}, and subjectivity \cite{rashkin2017truth}. For the purposes of this paper, we focus on investigating intervention effect on sentiment. for that we leverage DistilBERT model fine-tuned on  Stanford Sentiment Treebank (SST-2) provided by Huggingface. 
Sentiment analysis yields a quantitative measure, enabling us to assess the long-term impact of various mediation strategies. By comparing the average sentiment scores across generated conversations for different interventions, we can evaluate the sustained effectiveness of each approach in shaping the tone and emotional content of the discussions. Empathy is a categorical value that is divided into cognitive, understanding and interpreting a situation of another person (intent), and affective, the emotional reaction (emotion). 

\subsection{Experimental Setup}
Our experimental setup integrates multiple components to generate a comprehensive dataset for this study. We simulate conversations across three distinct platforms: Reddit, Twitter, and ED Support Forum. For each platform, we develop seven unique prompt types, reflecting the diverse communities and topics prevalent in each space. We implement seven mediation strategies, including an unmediated control condition, to assess varying levels of intervention. The conversation generation process employs seven different language models to ensure diversity in AI-generated responses. For each unique combination of platform, prompt type, mediation strategy, and language model, we generate 100 simulated conversations. Each conversation consists of 10 dialogue turns between two participants. This systematic approach yields a rich, multifaceted dataset that allows for robust analysis of intervention effectiveness across different social media platforms and LLMs.

 \begin{figure*}[t!]
 \centering
   \includegraphics[width=0.9\textwidth]{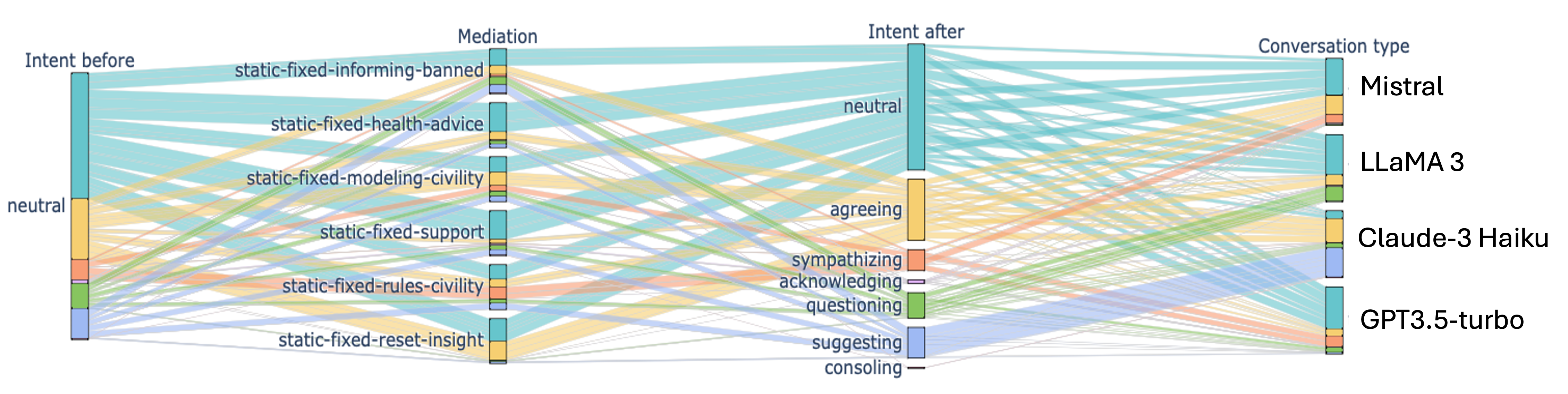}
  \caption{{\bf Cross-model analysis:} Comparing intervention effectiveness across different mediation strategies using
empathy intent metrics across LLM.}
   \label{fig:cross-model-empathy-intent}
   \vspace{-0.3cm}
 \end{figure*}

 \begin{figure*}[t!]
 \centering
   \includegraphics[width=0.9\textwidth]{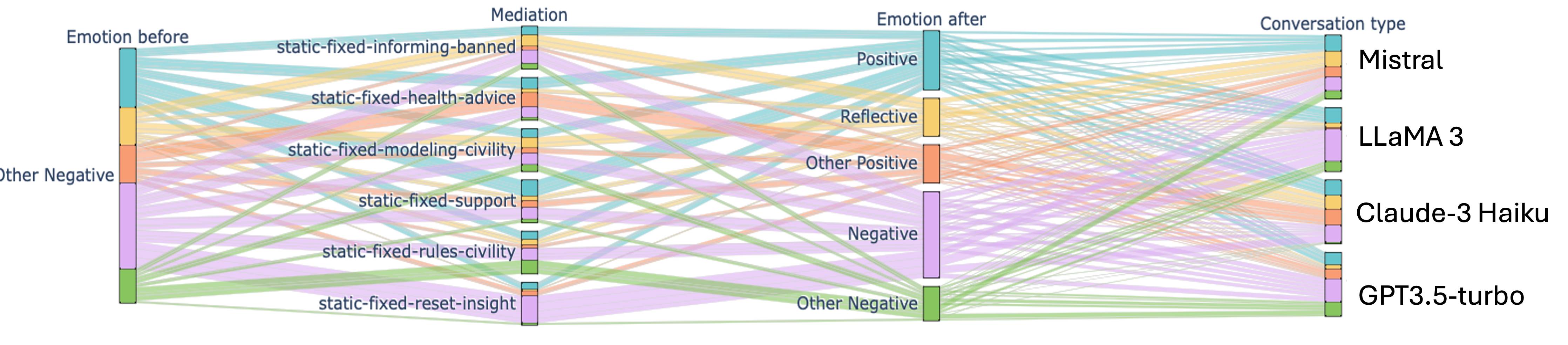}
   \caption{{\bf Cross-model analysis:} Comparing intervention effectiveness across different mediation strategies using
empathy emotion metrics across LLM.}
   \label{fig:cross-model-empathy-emotion}
      \vspace{-0.3cm}
 \end{figure*}

\section{Results and Discussion}
Our analysis of results is structured along three key dimensions, each isolating a specific variable while controlling for others:
\begin{itemize}
\item {\bf Cross-model:} Comparing different language models while keeping the platform and communities constant.
\item {\bf Cross-platform:} Evaluating different social media platforms while maintaining consistent models and communities.
\item {\bf Cross-community:} Examining  communities while controlling for model and platform.
\end{itemize}

\begin{figure*}[t!]
  \includegraphics[width=\textwidth]{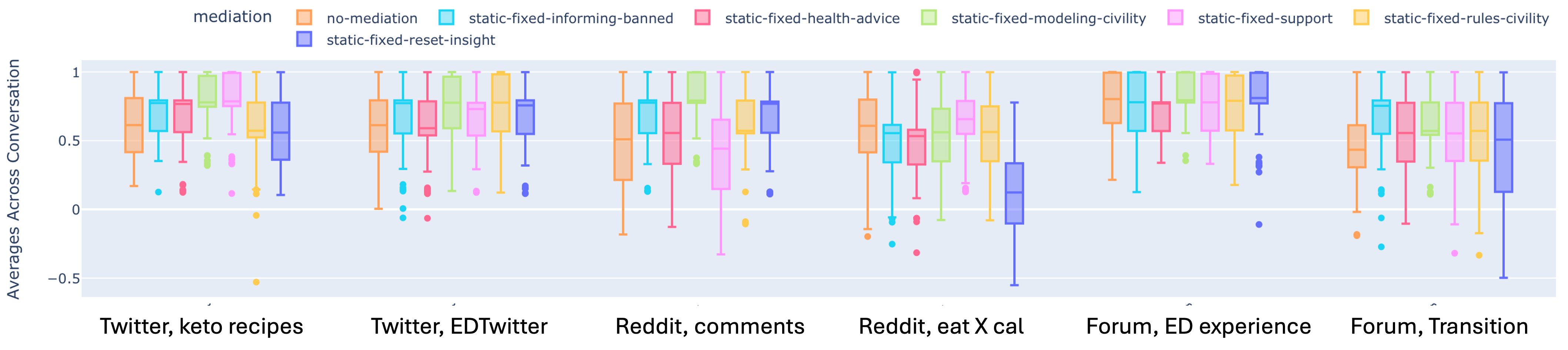}
  \caption{{\bf Cross-platform analysis:} Comparing intervention effectiveness across different mediation strategies using average sentiment scores over simulated conversations grouped by top 2 communities on each platform.}
  \label{fig:cross-platform-sentiment}
\end{figure*}

\begin{table*}[htb!]
\centering
\small
\begin{tabular}{| l | r | r | l | l |} 
 \hline
 Community & $Med(s_{none})$ & 
$\overline{Med(s)_{diff}}$ & $max(Med(s)_{diff})$ & $min(Med(s)_{diff})$   \\ [0.5ex] 
 \hline
 Twitter, keto recipes & 0.613 & 0.093 & 0.173, support & - 0.054, reset insight \\ 
 \hline
 Twitter, EDTwitter & 0.613 & 0.121 & 0.164, modeling civility & -0.022, health advice\\
 \hline
 Reddit, comments & 0.510 & 0.141 & 0.281, modeling civility & -0.069, support \\
 \hline
 Reddit, eat X cal & 0.608 & -0.109 & 0.049, support & -0.485, reset insight\\
 \hline
 ED Forum, ED exp & 0.803 & -0.015 & -0.031, reset insight & -0.092, health advice\\ 
 \hline
 ED Forum, transition & 0.434 & 0.151 & 0.320, informing banned & 0.073, reset insight\\ 
 \hline
\end{tabular}
\caption{Median, min and max sentiment scores across platforms with most and least effective interventions.}
\label{table:cross-platform-sentiment}
\vspace{-0.3cm}
\end{table*}

To visualize mediation effects, we employ box plots, focusing on median values to gauge sentiment in conversations. For clarity, we define sentiment as the average sentiment across all conversations for a given prompt, mediation, and model combination. We establish a baseline using the median sentiment of unmediated conversations ($s_{none}$). The impact of interventions is quantified by $Med(s)_{diff}$, which represents the difference between mediated and unmediated conversation sentiment medians. To assess overall mediation sensitivity, we calculate the average $Med(s)_{diff}$. Additionally, we consider the maximum and minimum $Med(s)_{diff}$ values to account for bidirectional sentiment shifts caused by mediations. This comprehensive approach allows for a nuanced understanding of how various factors influence the effectiveness of intervention strategies in online ED discussions.

Our analysis of empathy intent and emotion metrics provided additional insights into the nuanced effects of interventions. As shown in Figures~\ref{fig:cross-model-empathy-intent} and ~\ref{fig:cross-model-empathy-emotion}, empathy intent and emotion scores varied across different LLMs and intervention strategies. Supportive interventions generally increased expressions of empathy and positive intent in conversations. Civility-based mediations (i.e., discourse handling and informing community about rules violations) leads to most positive emotion, non-neutral intent, and on average increased sentiment across models. However, the relationship between interventions and emotional responses was complex, with some strategies reducing negative emotions without necessarily increasing positive ones. This highlights the need for a multifaceted approach to evaluating intervention effectiveness.

%

\subsection{Cross-Model Analysis}
We study the effect in the top community on Twitter and compare sentiment and empathy measures of discourse, across different LLMs. The average sentiment values of the conversations are visualized in box plots in Figure~\ref{fig:cross-model-sentiment}. Overall, we see the Claude 3 models are overwhelmingly more positive compared to the other generations, whereas the LLaMA3 generations have mostly negative sentiment. We show the numerical values of the metrics of interest in Table \ref{table:cross-model-sentiment}. The unmediated conversations have in order of median sentiment, from most positive to least positive: Claude-3 Haiku $>$ Mistral $>$ GPT3.5-turbo $>$ LLaMA3.


To identify the most impactful mediation, we also look at the $max(Med(s)_{diff})$ and $min(Med(s)_{diff})$, and find that for most models, {\it modeling civility} tended to lead to the most increase in sentiment, while {\it resetting insight} led to a decrease in sentiment. In addition, we investigated the effect of LLM versioning on mediation effectiveness. We found that GPT3.5 model became progressively more negative as new checkpoints were developed. These changes could be explained by an increase in generalizability of the model.

\begin{figure*}[t!]
  \includegraphics[width=\linewidth]{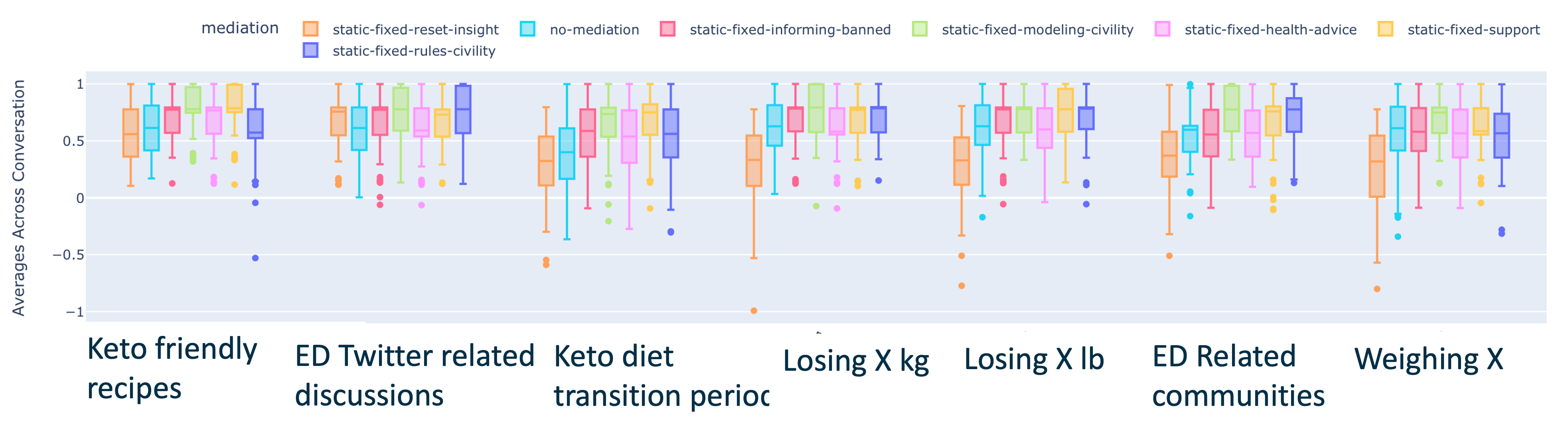}
  \caption{{\bf Cross-community analysis on Twitter:} Comparing intervention effectiveness across different mediation strategies across communities on Twitter using average sentiment scores over simulated conversations.}
  \label{fig:cross-community-sentiment}
\end{figure*}

\begin{table*}[htb!]
\centering
\small
\begin{tabular}{| l | r | r | l | l |} 
 \hline
 Community & $Med(s_{none})$ & $\overline{Med(s)_{diff}}$ & $max(Med(s)_{diff})$ & $min(Med(s)_{diff})$  \\ [0.5ex] 
 \hline
 keto recipes & 0.613 & 0.093 & 0.173, support & -0.054, reset insight \\ 
 \hline
 EDTwitter & 0.613 & 0.121 & 0.164, modeling civility & -0.022, health advice\\
 \hline
 Keto transition & 0.400 & 0.182 & 0.350, support & -0.078, reset insight \\
 \hline
 Losing X kg & 0.623 & 0.053 & 0.170, modeling civility & -0.291, reset insight\\
 \hline
 Losing X lb & 0.628 & 0.048 & 0.164, rules civility & -0.299, reset insight\\ 
 \hline
 ED-related & 0.598 & 0.037 & 0.178, modeling civility & -0.228, reset insight\\ 
 \hline
 Weighing X lb & 0.611 & -0.050 & 0.137, modeling civility & -0.292, reset insight\\  
 \hline
\end{tabular}
\caption{Median, min and max sentiments across communities on Twitter with most and least effective interventions.}
\label{table:cross-comunity-sentiment}
\vspace{-0.3cm}
\end{table*}

\subsection{Cross-Platform Analysis}
We study the effect of interventions across different platforms, by analyzing generations done by the Mistral model for  top two communities for each platform - Twitter, Reddit and ED Forum. The average sentiment values of the conversations are visualized using a box plot in Figure~\ref{fig:cross-platform-sentiment}, with aggregated metrics in Table \ref{table:cross-platform-sentiment}. We perform similar analysis as before, looking at $Med(s_{none})$, we see that the top two communities on Twitter and Reddit are relatively close to each other in terms of sentiment, whereas the ED Support forum has a wider range. We also find from $\overline{Med(s)_{diff}}$ that there are some community and platform combinations that encounter little to negative change from interventions, as in the case of the Reddit, {\it eat X cal} and {\it Forum, ED sharing communities}, and the opposite as well, as is the case of the {\it ED Forum, transition} and {\it Reddit, comments communities}. We find that {\it modeling civility} and {\it support} interventions are the most effective, whereas the {\it reset insight} and {\it health advice} are the least effective.

\subsection{Cross-Community Analysis}

To analyze the effects of mediation methods in different communities/topics in a platform, we focus on the generations of conversations in the Twitter platform by the Mistral model. The sentiment box plots are shown in Fig.~\ref{fig:cross-community-sentiment} and the metrics are displayed in Table \ref{table:cross-comunity-sentiment}. Outside of the keto transition community, the unmediated conversations have similar $Med(s_{none})$, but the impact of the mediations vary per community.  Reset insight mediation has been shown to drastically reduce sentiment across the board, especially in communities related to food estimates and losing weight (numbers). Modeling civility show to increase sentiment across the board. 

\section{Impact}
Our research demonstrates that the effectiveness of interventions varies across platforms, communities, and language models, highlighting the need for tailored approaches in military contexts where service members may engage in diverse online spaces. The discovery that civility-focused interventions consistently improve sentiment across different dimensions provides a valuable strategy for maintaining positive online environments \cite{kendal2017moderated,park2022topics,ransom2010interpersonal}. Conversely, the finding that insight-resetting approaches often lead to reduced sentiment and emotions cautions against their indiscriminate use. The identified biases in LLM-generated conversations, with perspectives varying notably across models and versions, underscores the importance of careful LLM selection and continuous evaluation in developing support tools for military personnel \cite{wan2023kelly,taubenfeld2024systematic,lin2024investigating}. These insights can directly inform the design of more effective online support systems and intervention strategies for service members, potentially mitigating the spread of harmful information related to eating disorders and other mental health challenges. 

\section{Conclusions}
Our study demonstrates that civility-focused interventions consistently improve sentiment across platforms and communities, while insight-resetting approaches often lead to negative outcomes. We demonstrated the potential of using LLM-driven simulations to generate and assess intervention strategies for sensitive online discussions. Our findings reveal that the effectiveness of interventions varies significantly across different models, platforms, and communities, highlighting the need for tailored approaches in online support systems. 



\section{Limitations}
While our study provides valuable insights into intervention strategies for ED-related discussions, it is important to acknowledge several limitations. The simulated environment, while allowing for controlled experimentation, may not fully capture the nuances and complexities of real-world online interactions. Inherent biases in the LLMs used may influence the generated conversations and subsequent analysis, potentially not accurately reflecting the diversity of real-world ED discussions~\cite{volkova2023explaining}. Our focus on three specific platforms and a set number of ED-related topics may not encompass the full range of ED discussions occurring across various online spaces~\cite{saldanha2019evaluation}. Additionally, our analysis provides a snapshot of intervention effectiveness but does not account for long-term effects or changes in online discourse over time. The lack of direct input from individuals affected by EDs limits our ability to fully assess the perceived helpfulness or potential negative impacts of the tested interventions. Furthermore, the effectiveness of interventions may vary in real-world applications due to factors not captured in our simulated environment. These limitations underscore the need for cautious interpretation of our results and highlight areas for future research to further validate and expand upon our findings.

Future work will focus on: (1) investigating long-term intervention effects through extended simulations and longitudinal analysis and (2) incorporating multimodal data to improve simulation realism.

\section{Ethical Considerations}
The highly sensitive nature of eating disorder discussions and the potential impact of interventions on vulnerable individuals underscore the importance of testing intervention strategies in a controlled, simulated environment before real-world application. This approach forms the ethical foundation of our research, allowing us to explore and refine intervention techniques without risking harm to individuals struggling with EDs. In developing and conducting our study, we have taken several steps to ensure the ethical integrity of our research:
\begin{itemize}[noitemsep, nolistsep]
\item {\it Simulation-Based Testing:} By using LLM-generated conversations, we avoid direct interaction with vulnerable individuals, minimizing potential harm while still gaining valuable insights into intervention strategies. This simulated environment allows us to test a wide range of interventions and assess their potential impacts without exposing real individuals to untested approaches.
\item {\it Data Privacy and Protection:} In our use of data from ED Support Forum, we have rigorously removed all personally identifiable information (PII) and sensitive details to protect user privacy. This approach allows us to analyze authentic ED-related discussions while maintaining strict adherence to ethical research standards.
\item {\it Diverse Representation:} Our use of multiple platforms (Twitter, Reddit, ED Support Forum) ensures a broad representation of ED-related discussions, reducing bias that might result from focusing on a single platform or community.
\item {\it Transparency in LLM Limitations:} We acknowledge and report on the biases and limitations of the LLMs used in our study, ensuring transparency in our methodology and results.
\item {\it Interdisciplinary Approach:} Our intervention strategies are informed by principles of civil discourse and clinical expertise, ensuring a holistic approach to addressing ED-related discussions.
\item {\it Potential for Positive Impact:} While our study does not directly intervene in real-world discussions, it aims to inform more effective and sensitive intervention strategies, potentially benefiting individuals affected by EDs in the long term.
\item {\it Continuous Evaluation:} We emphasize the need for ongoing evaluation and refinement of intervention strategies, acknowledging that the digital landscape and the nature of ED discussions are continually evolving.
\item {\it Expert Validation:} We recommend that our findings be further validated by subject matter experts before being applied in real-world scenarios, ensuring responsible use of our research outcomes.
\end{itemize}

By adhering to these ethical principles and prioritizing a simulation-based approach, we strive to conduct research that not only advances our understanding of ED-related online discussions but also respects the dignity and well-being of the individuals these discussions represent. This methodology allows us to explore potentially beneficial interventions while minimizing the risk of unintended negative consequences in real-world applications.

\bibliography{acl_latex}

\begin{thebibliography}{32}
\providecommand{\natexlab}[1]{#1}

\bibitem[{{American Psychiatry Assoication}(2023)}]{psy2023us}
{American Psychiatry Assoication}. 2023.
\newblock \href {https://www.psychiatry.org/patients-families/eating-disorders/what-are-eating-disorders} {What are eating disorders?}
\newblock \emph{Eating Disorders}.

\bibitem[{Antczak and Brininger(2008)}]{antczak2008diagnosed}
Amanda~J Antczak and Teresa~L Brininger. 2008.
\newblock Diagnosed eating disorders in the us military: A nine year review.
\newblock \emph{Eating Disorders}, 16(5):363--377.

\bibitem[{Bauman et~al.(2024)Bauman, Thompson, Sunderland, Thornton, Schvey, Sekyere, Funk, Pav, Brydum, Klein et~al.}]{bauman2024incidence}
Viviana Bauman, Katherine~A Thompson, Kevin~W Sunderland, Jennifer~A Thornton, Natasha~A Schvey, Nana~Amma Sekyere, Wendy Funk, Veronika Pav, Rick Brydum, David~A Klein, et~al. 2024.
\newblock Incidence and prevalence of eating disorders among us military service members, 2016--2021.
\newblock \emph{International Journal of Eating Disorders}.

\bibitem[{Bodell et~al.(2014)Bodell, Forney, Keel, Gutierrez, and Joiner}]{bodell2014consequences}
Lindsay Bodell, Katherine~Jean Forney, Pamela Keel, Peter Gutierrez, and Thomas~E Joiner. 2014.
\newblock Consequences of making weight: A review of eating disorder symptoms and diagnoses in the u nited s tates military.
\newblock \emph{Clinical Psychology: Science and Practice}, 21(4):398--409.

\bibitem[{{Centers for Disease Control and Prevention}(2022)}]{ped2023us}
{Centers for Disease Control and Prevention}. 2022.
\newblock \href {https://www.cdc.gov/mmwr/volumes/71/wr/mm7108e2.htm} {Pediatric emergency department visits associated with mental health conditions before and during the covid 19 pandemic united states, january 2019 to january 2022}.
\newblock \emph{Morbidity and Mortality Weekly Report}.

\bibitem[{Devlin et~al.(2018)Devlin, Chang, Lee, and Toutanova}]{DBLP:journals/corr/abs-1810-04805}
Jacob Devlin, Ming{-}Wei Chang, Kenton Lee, and Kristina Toutanova. 2018.
\newblock \href {https://arxiv.org/abs/1810.04805} {{BERT:} pre-training of deep bidirectional transformers for language understanding}.
\newblock \emph{CoRR}, abs/1810.04805.

\bibitem[{Garten et~al.(2016)Garten, Boghrati, Hoover, Johnson, and Dehghani}]{garten2016morality}
Justin Garten, Reihane Boghrati, Joe Hoover, Kate~M Johnson, and Morteza Dehghani. 2016.
\newblock Morality between the lines: Detecting moral sentiment in text.
\newblock In \emph{Proceedings of IJCAI 2016 workshop on Computational Modeling of Attitudes}.

\bibitem[{Hanu and {Unitary team}(2020)}]{Detoxify}
Laura Hanu and {Unitary team}. 2020.
\newblock Detoxify.
\newblock Github. https://github.com/unitaryai/detoxify.

\bibitem[{Huggingface()}]{sentiment:huggingface}
Huggingface.
\newblock Distilbert base uncased finetuned sst-2 model.
\newblock \url{https://huggingface.co/distilbert/distilbert-base-uncased-finetuned-sst-2-english}.

\bibitem[{Kao et~al.(2024)Kao, Erickson, Chu, He, Lerman, and Volkova}]{Kao_CHI_24}
Hsien-Te Kao, Isabel Erickson, Minh Duc~Hoang Chu, Zihao He, Kristina Lerman, and Svitlana Volkova. 2024.
\newblock \href {https://doi.org/10.1145/3613905.3651116} {Machine learning insights into eating disorder twitter communities}.
\newblock In \emph{Extended Abstracts of the 2024 CHI Conference on Human Factors in Computing Systems}, CHI EA '24, New York, NY, USA. Association for Computing Machinery.

\bibitem[{Kendal et~al.(2017)Kendal, Kirk, Elvey, Catchpole, and Pryjmachuk}]{kendal2017moderated}
Sarah Kendal, Sue Kirk, Rebecca Elvey, Roger Catchpole, and Steven Pryjmachuk. 2017.
\newblock How a moderated online discussion forum facilitates support for young people with eating disorders.
\newblock \emph{Health Expectations}, 20(1):98--111.

\bibitem[{Lee et~al.(2022)Lee, Lim, and Choi}]{lee2022does}
Young-Jun Lee, Chae-Gyun Lim, and Ho-Jin Choi. 2022.
\newblock Does gpt-3 generate empathetic dialogues? a novel in-context example selection method and automatic evaluation metric for empathetic dialogue generation.
\newblock In \emph{Proceedings of the 29th International Conference on Computational Linguistics}, pages 669--683.

\bibitem[{Lerman et~al.(2023)Lerman, Karnati, Zhou, Chen, Kumar, He, Yau, and Horn}]{twitter2023}
Kristina Lerman, Aryan Karnati, Shuchan Zhou, Siyi Chen, Sudesh Kumar, Zihao He, Joanna Yau, and Abigail Horn. 2023.
\newblock \href {https://arxiv.org/abs/2305.11316} {Radicalized by thinness: Using a model of radicalization to understand pro-anorexia communities on twitter}.
\newblock \emph{Preprint}, arXiv:2305.11316.

\bibitem[{Lin et~al.(2024)Lin, Wang, Guo, and Wong}]{lin2024investigating}
Luyang Lin, Lingzhi Wang, Jinsong Guo, and Kam-Fai Wong. 2024.
\newblock Investigating bias in llm-based bias detection: Disparities between llms and human perception.
\newblock \emph{arXiv preprint arXiv:2403.14896}.

\bibitem[{{Mayo Clinic}(2023)}]{mayo2023us}
{Mayo Clinic}. 2023.
\newblock \href {https://www.mayoclinic.org/diseases-conditions/eating-disorders/symptoms-causes/syc-20353603} {Eating disorders}.
\newblock \emph{Diseases and Conditions}.

\bibitem[{Mitchell et~al.(2023)Mitchell, Smith, Masheb, and Vogt}]{mitchell2023impact}
Karen~S Mitchell, Brian~N Smith, Robin Masheb, and Dawne Vogt. 2023.
\newblock The impact of the covid-19 pandemic on eating disorders in us military veterans.
\newblock \emph{International Journal of Eating Disorders}, 56(1):108--117.

\bibitem[{{National Institute of Mental Health}(2023)}]{nimh2023us}
{National Institute of Mental Health}. 2023.
\newblock \href {https://www.nimh.nih.gov/health/topics/eating-disorders} {Eating disorders}.
\newblock \emph{Health Topics}.

\bibitem[{Park et~al.(2022)Park, Kim, and Kim}]{park2022topics}
Eunhye Park, Woo-Hyuk Kim, and Sung-Bum Kim. 2022.
\newblock What topics do members of the eating disorder online community discuss and empathize with? an application of big data analytics.
\newblock In \emph{Healthcare}, volume~10, page 928. MDPI.

\bibitem[{Ransom et~al.(2010)Ransom, La~Guardia, Woody, and Boyd}]{ransom2010interpersonal}
Danielle~C Ransom, Jennifer~G La~Guardia, Erik~Z Woody, and Jennifer~L Boyd. 2010.
\newblock Interpersonal interactions on online forums addressing eating concerns.
\newblock \emph{International Journal of Eating Disorders}, 43(2):161--170.

\bibitem[{Rashkin et~al.(2017)Rashkin, Choi, Jang, Volkova, and Choi}]{rashkin2017truth}
Hannah Rashkin, Eunsol Choi, Jin~Yea Jang, Svitlana Volkova, and Yejin Choi. 2017.
\newblock Truth of varying shades: Analyzing language in fake news and political fact-checking.
\newblock In \emph{Proceedings of the 2017 conference on empirical methods in natural language processing}, pages 2931--2937.

\bibitem[{Rashkin et~al.(2015)Rashkin, Singh, and Choi}]{rashkin2015connotation}
Hannah Rashkin, Sameer Singh, and Yejin Choi. 2015.
\newblock Connotation frames: A data-driven investigation.
\newblock \emph{arXiv preprint arXiv:1506.02739}.

\bibitem[{Reimers and Gurevych(2019)}]{reimers2019sentence}
Nils Reimers and Iryna Gurevych. 2019.
\newblock Sentence-bert: Sentence embeddings using siamese bert-networks.
\newblock \emph{arXiv preprint arXiv:1908.10084}.

\bibitem[{Saldanha et~al.(2019)Saldanha, Blaha, Sathanur, Hodas, Volkova, and Greaves}]{saldanha2019evaluation}
Emily Saldanha, Leslie~M Blaha, Arun~V Sathanur, Nathan Hodas, Svitlana Volkova, and Mark Greaves. 2019.
\newblock Evaluation and validation approaches for simulation of social behavior: challenges and opportunities.
\newblock \emph{Social-Behavioral Modeling for Complex Systems}, pages 495--519.

\bibitem[{Smith et~al.(2020)Smith, Emerson, Winkelmann, Potter, and Torres-McGehee}]{smith2020prevalence}
Allison Smith, Dawn Emerson, Zachary Winkelmann, Devin Potter, and Toni Torres-McGehee. 2020.
\newblock Prevalence of eating disorder risk and body image dissatisfaction among rotc cadets.
\newblock \emph{International Journal of Environmental Research and Public Health}, 17(21):8137.

\bibitem[{Sánchez et~al.(2024)Sánchez, Chu, He, Dorn, Murray, and Lerman}]{reddit2024}
Cinthia Sánchez, Minh~Duc Chu, Zihao He, Rebecca Dorn, Stuart Murray, and Kristina Lerman. 2024.
\newblock \href {https://arxiv.org/abs/2407.03551} {Feelings about bodies: Emotions on diet and fitness forums reveal gendered stereotypes and body image concerns}.
\newblock \emph{Preprint}, arXiv:2407.03551.

\bibitem[{Taubenfeld et~al.(2024)Taubenfeld, Dover, Reichart, and Goldstein}]{taubenfeld2024systematic}
Amir Taubenfeld, Yaniv Dover, Roi Reichart, and Ariel Goldstein. 2024.
\newblock Systematic biases in llm simulations of debates.
\newblock \emph{arXiv preprint arXiv:2402.04049}.

\bibitem[{{The White House}(2023)}]{whitehouse2023us}
{The White House}. 2023.
\newblock \href {https://www.whitehouse.gov/briefing-room/statements-releases/2023/05/23/fact-sheet-biden-harris-administration-announces-actions-to-protect-youth-mental-health-safety-privacy-online/} {Fact sheet: Biden harris administration announces actions to protect youth mental health, safety, and privacy online}.
\newblock \emph{Briefing Room}.

\bibitem[{{US Department of Health and Human Services}(2023)}]{hhs2023us}
{US Department of Health and Human Services}. 2023.
\newblock \href {https://www.hhs.gov/about/news/2023/05/23/surgeon-general-issues-new-advisory-about-effects-social-media-use-has-youth-mental-health.html} {Surgeon general issues new advisory about effects social media use has on youth mental health}.
\newblock \emph{News}.

\bibitem[{{US Senate Committee on Health, Education, Labor, and Pensions}(2023)}]{senate2023us}
{US Senate Committee on Health, Education, Labor, and Pensions}. 2023.
\newblock \href {https://www.help.senate.gov/hearings/why-are-so-many-american-youth-in-a-mental-health-crisis-exploring-causes-and-solutions} {Why are so many american youth in a mental health crisis? exploring causes and solutions}.
\newblock \emph{Full Committee Hearing}.

\bibitem[{van Hoeken and Hoek(2020)}]{Hoeken2020}
Daphne van Hoeken and Hans Hoek. 2020.
\newblock \href {https://doi.org/doi:10.1097/YCO.0000000000000641} {Review of the burden of eating disorders: mortality, disability, costs, quality of life, and family burden}.
\newblock \emph{Current opinion in psychiatry}, 33, 6:521--527.

\bibitem[{Volkova et~al.(2023)Volkova, Arendt, Saldanha, Glenski, Ayton, Cottam, Aksoy, Jefferson, and Shrivaram}]{volkova2023explaining}
Svitlana Volkova, Dustin Arendt, Emily Saldanha, Maria Glenski, Ellyn Ayton, Joseph Cottam, Sinan Aksoy, Brett Jefferson, and Karthnik Shrivaram. 2023.
\newblock Explaining and predicting human behavior and social dynamics in simulated virtual worlds: reproducibility, generalizability, and robustness of causal discovery methods.
\newblock \emph{Computational and Mathematical Organization Theory}, 29(1):220--241.

\bibitem[{Wan et~al.(2023)Wan, Pu, Sun, Garimella, Chang, and Peng}]{wan2023kelly}
Yixin Wan, George Pu, Jiao Sun, Aparna Garimella, Kai-Wei Chang, and Nanyun Peng. 2023.
\newblock " kelly is a warm person, joseph is a role model": Gender biases in llm-generated reference letters.
\newblock \emph{arXiv preprint arXiv:2310.09219}.

\end{thebibliography}

\end{document}